\pdfoutput=1

\documentclass[11pt]{article}

\usepackage[final]{acl}

\usepackage{times}
\usepackage{latexsym}

\usepackage[T1]{fontenc}

\usepackage[utf8]{inputenc}

\usepackage{microtype}

\usepackage{inconsolata}

\usepackage{graphicx}

\usepackage{CJKutf8}
\usepackage{tabularx}

\title{Word Order in English-Japanese Simultaneous Interpretation: \\Analyses and Evaluation using Chunk-wise Monotonic Translation}

\author{
 \textbf{Kosuke Doi\textsuperscript{1}},
 \textbf{Yuka Ko\textsuperscript{1}},
 \textbf{Mana Makinae\textsuperscript{1}},
 \textbf{Katsuhito Sudoh\textsuperscript{1, 2}},
 \textbf{Satoshi Nakamura\textsuperscript{1, 3}}
\\
\\
 \textsuperscript{1}Nara Institute of Science and Technology, \\
 \textsuperscript{2}Nara Women's University,
 \textsuperscript{3}The Chinese University of Hong Kong, Shenzhen
\\
 \small{
   \texttt{\{doi.kosuke.de8, sudoh, s-nakamura\}@is.naist.jp}
 }
}

\begin{document}
\maketitle
\begin{abstract}
This paper analyzes the features of monotonic translations, which follow the word order of the source language, in simultaneous interpreting (SI).
Word order differences are one of the biggest challenges in SI, especially for language pairs with significant structural differences like English and Japanese.
We analyzed the characteristics of chunk-wise monotonic translation (CMT) sentences using the NAIST English-to-Japanese Chunk-wise Monotonic Translation Evaluation Dataset and identified some grammatical structures that make monotonic translation difficult in English-Japanese SI.
We further investigated the features of CMT sentences by evaluating the output from the existing speech translation (ST) and simultaneous speech translation (simulST) models on the NAIST English-to-Japanese Chunk-wise Monotonic Translation Evaluation Dataset as well as on existing test sets.
The results indicate the possibility that the existing SI-based test set underestimates the model performance.
The results also suggest that using CMT sentences as references gives higher scores to simulST models than ST models, and that using an offline-based test set to evaluate the simulST models underestimates the model performance.
\end{abstract}

\section{Introduction}

Simultaneous interpreting (SI) is the task of translating speech from a source language into a target language in real time.
SI is cognitively demanding, and human simultaneous interpreters employ such strategies as segmentation, summarization, and generalization \cite{he-etal-2016-interpretese}.
Maintaining word order in a source language is another important strategy, especially for language pairs whose word order differs (\emph{e.g.}, English and Japanese), to shorten delays and reduce cognitive load.
Because of these features, SI sentences are different from offline translation sentences, although most automatic SI studies \cite{oda-etal-2014-optimizing, ma-etal-2019-stacl, liu20s_interspeech, papi23_interspeech} have used offline translation corpora (\emph{e.g.}, MuST-C; \citealp{di-gangi-etal-2019-must}) for both training and evaluatng models due to the limited amount of simultaneous interpretation corpora (SICs).

For English-Japanese language pairs, several SICs have been constructed \cite{tohyama-et-al-2004,shimizu-etal-2014-collection,matsushita-et-al-2020,doi-etal-2021-large}.
Based on the NAIST Simultaneous Interpretation Corpus (NAIST-SIC; \citealp{doi-etal-2021-large}), \citet{zhao-2024-naist-sic-aligned}\footnote{The dataset was released in 2023 (see version 3  of the paper).} created an automatically-aligned parallel SI dataset: NAIST-SIC-Aligned.
Since its sentences are aligned at the sentence level, they can be used for model training.
Actually, \citet{ko-etal-2023-tagged} and \citet{zhao-2024-naist-sic-aligned} trained SI models using SI data from NAIST-SIC-Aligned.
Their model performances were evaluated through automatic evaluation metrics such as BLEU \cite{papineni-etal-2002-bleu} using a small test set curated based on SI sentences generated by professional human simultaneous interpreters.

Although the scores reported in \citet{ko-etal-2023-tagged} and \citet{zhao-2024-naist-sic-aligned} were relatively low, the test set used in both studies might have underestimated the model performance.
Since human simultaneous interpreters use such strategies as summarization and generalization, phrases that do not affect the main idea are not necessarily translated into the target language.
If an SI model generates translations for phrases that a human interpreter did not, the output sentence might not be evaluated properly, even when it is a \emph{correct} translation.

\begin{table*}
\centering
\small
\begin{tabularx}{\linewidth}{lX}
\hline
Source & (1) The US Secret Service, / (2) two months ago, / (3) froze the Swiss bank account / (4) of Mr. Sam Jain right here, / (5) and that bank account / (6) had 14.9 million US dollars in it / (7) when it was frozen. \\
\hline
Offline & \begin{CJK}{UTF8}{min} {\small (1) 米国のシークレットサービスは / (2) 2ヶ月前に / (4) サム・ジェイン氏の / (3) スイス銀行口座を凍結しました / (5) その口座には / (6) 米ドルで1490万ドルありました} \end{CJK} \\
 & [The US Secret Service / two months ago / Mr. Sam Jain's / froze the Swiss bank account / that bank account / had 14.9 million US dollars] \\
\hline
SI & \begin{CJK}{UTF8}{min} {\small (1) アメリカのシークレッドサービスが、/ (3) スイスの銀行の口座を凍結しました。 / (4) サムジェインのものです。 / (5) この銀行口座の中には、 / (6) 一千四百九十万ドルが入っていました。} \end{CJK} \\
 & [The US Secret Service / froze the Swiss bank account / it is Sam Jain's one / in this bank account / had 14.9 million dollars] \\
\hline
CMT & \begin{CJK}{UTF8}{min} {\small (1) アメリカ合衆国シークレットサービスは、 / (2) 2ヶ月前に、 / (3) スイスの銀行口座を凍結しました、 / (4) ここにいるサム・ジェイン氏の口座です、 / (5) そしてその銀行口座には / (6) 490万米ドルが入っていました、 / (7) 凍結された時。} \end{CJK} \\
 & [The US Secret Service / two months ago / froze the Swiss bank account / the account of Mr. Sam Jain right here / and that bank account / had 14.9 million US dollars in it / when it was frozen] \\
\hline
\end{tabularx}
\caption{Comparison of target sentences in each translation mode. Examples of offline, SI, and CMT are respectively from subtitles of TED talks, NAIST-SIC, and NAIST English-to-Japanese Chunk-wise Monotonic Translation Evaluation Dataset. ``/'' shows boundaries of chunks. Numbers preceding chunks in source sentence represent appearance order. Numbers preceding chunks in target sentences correspond to numbers in source sentence.}
\label{tab:comparison-of-translation-mode}
\end{table*}

\citet{Fukuda2024-cwmt} pointed out the difficulty for SI models to learn which phrases in source speech are less important and advocated constructing SI models that only employ a strategy that maintains the word order in a source language.
As a first step, they created the NAIST English-to-Japanese Chunk-wise Monotonic Translation Evaluation Dataset\footnote{\url{https://dsc-nlp.naist.jp/data/NAIST-SIC/Aligned-Chunk_Mono-EJ/}}.
The source sentences in the test set used in \citet{ko-etal-2023-tagged} were automatically segmented into chunks, each of which was translated in a way that did not include the content of subsequent chunks.
Unlike in SI sentences by human interpreters, where not all the information in the source sentences is translated, chunk-wise monotonic translation (CMT) sentences\footnote{CMT refers to the task of segmenting a source sentence into chunks and translating it in the order of the chunks. A CMT sentence is a target sentence generated through CMT.} were translated so that all the information is translated (Table~\ref{tab:comparison-of-translation-mode})\footnote{Precisely, omissions that maintained the fluency of the sentence were allowed. See Section~\ref{ssec:Data} for the details about the dataset.}.
\citet{Fukuda2024-cwmt} have investigated the quality of the CMT sentences in their dataset through human evaluation, although they have not analyzed its characteristics.
Nor have they conducted any evaluation experiments in which model outputs are evaluated on their dataset.

In this paper, we qualitatively and quantitatively analyze CMT sentences in the NAIST English-to-Japanese Chunk-wise Monotonic Translation Evaluation Dataset.
In the process of generating CMT sentences for the dataset, it was allowed to \texttt{repeat}, \texttt{defer}, and \texttt{omit} phrases in the source sentences to maintain the translation's fluency.
We assume the presence of factors (\emph{e.g.}, syntactic structures) that prevent monotonic translation if phrases were \texttt{repeated}, \texttt{deferred}, or \texttt{omitted} in the CMT sentences since they were translated without time constraints.
In addition, we evaluate the output from an existing speech translation (ST) model and two simultaneous speech translation (simulST) models (See \ref{ssec:st-model}).
Both the ST\footnote{A ST model generates translations after the utterances are completed.} and simulST models are used to investigate the differences in scores when evaluating translations with different characteristics.
The contributions of this paper are as follows:
\begin{itemize}
    \item We analyze CMT sentences and show that they tend to be longer than offline translations primarily because of \texttt{repetition}.
    \item We investigate what causes the phrases in source sentences to be \texttt{repeated}, \texttt{deferred}, and \texttt{omitted} and show that most cases occur because of particular grammatical structures. When a phrase in a chunk is a dependent of a phrase in the preceding chunk, the head phrase is typically \texttt{repeated} or \texttt{deferred}.
    \item We evaluate the output from three different models on the NAIST English-to-Japanese Chunk-wise Monotonic Translation Evaluation Dataset: (1)~an ST model trained on offline data, (2)~a simulST model trained on offline data, and (3)~a simulST model trained on both offline and SI data.
    The results suggest that the existing SI-based test set \cite{ko-etal-2023-tagged,zhao-2024-naist-sic-aligned} underestimates the model performance. 
    The results also suggest that using CMT sentences as references gives higher scores to simulST models than ST models,
    while using an offline-based test set for evaluating simulST models underestimates the model performance.
\end{itemize}

\section{Related Work}

\subsection{Simultaneous Interpretation Corpora}

SICs are valuable resources both for developing automatic SI models and analyzing SI's characteristics.
For English-Japanese language pairs, several SICs are publicly available \cite{tohyama-et-al-2004,shimizu-etal-2014-collection,matsushita-et-al-2020,doi-etal-2021-large}, although the amount of such corpora is very limited compared to offline translation corpora.

Using these corpora, SI sentences have been analyzed from various perspectives, such as strategies and interpreting patterns used by interpreters, latency, translation quality, and word order \cite{tohyama-matsubara-2006-collection, ono-etal-2008-construction, cai-etal-2018-statistical, cai-2020-affects, doi-etal-2021-large}.
SI models have also been developed using SICs \cite{ryu-2004, shimizu-etal-2013-constructing, ko-etal-2023-tagged}.

\subsection{Word Order in Simultaneous Interpreting}

When dealing with language pairs whose sentence structures are different, including English and Japanese (SVO/head-initial vs. SOV/head-final), reducing the word order differences between the source and the target languages is crucial for minimizing delays.

\citet{murata-etal-2010-construction} segmented source sentences into semantically meaningful units with a maximum length of 4.3 seconds and translated those units from an SI viewpoint.
\citet{he-etal-2015-syntax} designed syntactic transformation rules for Japanese-English simultaneous machine translation.
By applying the rules to target language sentences (\emph{i.e.}, English), they generated more monotonic translations, while preserving the meaning of source sentences and maintaining the grammaticality of the target language.
In English-Japanese SI, \citet{futanata-2020} reordered Japanese sentences to make the word order closer to the original English sentences.
They further applied style transfer to increase the fluency and obtained sentences close to SI sentences by human interpreters.
\citet{han-etal-2021-monotonic} proposed an algorithm to reorder and refine the target sentences so that the target sentences were aligned largely monotonically.
They trained SI models for four language pairs, including English-Japanese.
\citet{nakabayashi-2021} segmented sentences into chunks and created bilingual pairs of such chunks with explicit annotations of context information.
The SI model trained on the data translated the source sentences while referencing the preceding chunks although naturally connecting chunks remained a challenge.
\citet{higashiyama-2023} constructed a large-scale English $\leftrightarrow$ Japanese SIC with the information of chunk boundaries in source and target sentences and phrases that can be omitted in target sentences.
The NAIST English-to-Japanese Chunk-wise Monotonic Translation Evaluation Dataset \cite{Fukuda2024-cwmt}, which is similar to \citeposs{higashiyama-2023}, is relatively small and intended for the evaluation purposes.

The word order differences among different translation modes have also been investigated.
\citet{okamura-2023} quantitatively compared the degree to which the word order of the source sentences was maintained and found that SI sentences retained the order better than consecutive interpreting and offline translation sentences.
\citet{cai-2020-affects} found syntactic and non-syntactic factors that affect interpreters' word order decisions through the statistical analyses of an SIC.
In this paper, we analyze what makes monotonic translation difficult.
While \citet{cai-2020-affects} analyzed the actual SI data generated by human simultaneous interpreters, we use CMT sentences, which were generated without time constraints, and in which all the information in the source sentences is translated into target sentences.

\section{Chunk-wise Monotonic Translation}

In SI between language pairs with different sentence structures, interpreters segment source sentences into chunks and translate them from chunk to chunk\footnote{Reordering may occur within a chunk.} \cite{okamura-2023}.
This section describes the details of the NAIST English-to-Japanese Chunk-wise Monotonic Translation Evaluation Dataset, which is used in our analyses.

\subsection{Data}
\label{ssec:Data}

The NAIST English-to-Japanese Chunk-wise Monotonic Translation Evaluation Dataset consists of 511 pairs of source sentences and their corresponding chunk-wise monotonic translation (CMT) sentences with information of chunk boundaries in the source and target sentences.
Source (\emph{i.e.}, English) sentences, which were used as the test set in \citet{ko-etal-2023-tagged}, were segmented following the five rules that reflected the interpreter's strategies\footnote{See the original paper or the code for chunking: \url{https://github.com/ahclab/si_chunker}} based on the syntactic analysis results from spaCy.
The source sentences come from eight TED talks.

\begin{table}
\centering
\begin{tabular}{lrr}
\hline
\textbf{Data} & \textbf{Sum} & \textbf{Per Sent.$\pm$SD} \\
\hline
\# sentence pairs  & 511 & -- \\
\# chunks & 1,677 & 3.28{$\pm$2.12} \\
\hline
\# source words & 8,104 & 15.86{$\pm$10.16} \\
\# target words & 13,981 & 27.36{$\pm$18.55} \\
\hline
\end{tabular}
\caption{Statistics for chunk-wise monotonic translation data. Standard deviations and number of words in target sentences were calculated by us. Other values are cited from \citet{Fukuda2024-cwmt}.}
  \label{tab:statistics-cwmt}
\end{table}

Translators were provided with source sentences with chunk boundaries and asked to (1)~translate them in the order of the chunks while (2)~naturally connecting the chunks and (3)~not including the content of subsequent chunks.
They were allowed to (1)~\texttt{repeat}, (2)~\texttt{defer}, and (3)~\texttt{omit} phrases in the source sentences to keep translation fluency, although they were instructed to minimize their use of the operations with larger number as much as possible (\emph{e.g.}, \texttt{defer} should not be used when \texttt{repeat} can handle the situation) .
Data statistics are shown in Table~\ref{tab:statistics-cwmt}.

\section{Data Analysis}

\citet{Fukuda2024-cwmt} have examined the quality of CMT sentences through human evaluation but have not analyzed the characteristics of them.
We suppose that factors exist that prevent monotonic translation if a phrase in the source sentences is \texttt{repeated}, \texttt{deferred}, or \texttt{omitted} since the CMT sentences were generated without time constraints.
Therefore we qualitatively and quantitatively analyze the CMT sentences with these operations and reveal such factors.

To better understand the characteristics of the CMT sentences, we also compare them with SI sentences from NAIST-SIC and NAIST-SIC-Aligned as well as offline translation sentences from the subtitles of TED talks.
Since the SI sentences in NAIST-SIC were not aligned at the sentence level, we manually align them.
Some source sentences did not match across the datasets, and we excluded those ten sentences from the analyses.
In addition, 25 sentences were not translated in NAIST-SIC\footnote{For example, due to time constraints, interpreters might have been unable to translate a whole sentence. See \citet{doi-etal-2021-large}.}, which were also excluded from the analyses.
As a result, 476 sentences were used for our analyses.

\begin{table*}
\centering
\small
\begin{tabularx}{\linewidth}{llX}
\hline
\textbf{Type} & \textbf{Tag} & \textbf{Meaning} \\
\hline
Span & repeat & Phrases that are repeated\\
 & zero-repeat & Target phrases that are repeated; No corresponding source phrases (\emph{e.g.}, zero that-clause) \\
 & defer & Phrases that are not translated within the current chunk but in a subsequent chunk \\
 & omit & Source phrases that are not translated in the target sentences \\
 & ahead & Target phrases translated using the subsequent chunks \\
 & add & Target phrases that have no corresponding source phrases  \\
 & error & Phrases with translation errors \\
 & sep & Boundaries of source and target sentences \\
\hline
Relation & rel-repeat & Connect source and target phrases with \texttt{repeat} tags \\
 & repeat\_d\# & Connect target phrases with \texttt{repeat} tags \\
 & defer\_d\# & Connect source and target phrases with \texttt{defer} tags \\
 & ahead\_d\# & Connect source and target phrases with \texttt{ahead} tags\\
 & rel-err & Connect source and target phrases with \texttt{error} tags \\
\hline
\end{tabularx}
\caption{List of tags used for annotations. ``d\#'' (\#=1, 2, ...) represents distance between chunks.}
  \label{tab:tag-list}
\end{table*}

\subsection{Annotations}

To analyze the characteristics of the CMT sentences, we annotated tags to the source and CMT sentences.
The list of tags is shown in Table~\ref{tab:tag-list}.
Prior to the annotations, we tokenized the English sentences using spaCy\footnote{\url{https://spacy.io/}} and the Japanese sentences using MeCab \cite{kudo-etal-2004-applying} with unidic.
Then, we concatenated the source and CMT sentences with a special token \texttt{[SEP]} and annotated them using an open-source data labeling tool, doccano.\footnote{\url{https://github.com/doccano/doccano}}

\begin{figure*}[t]
\centering
\includegraphics[width=13cm]{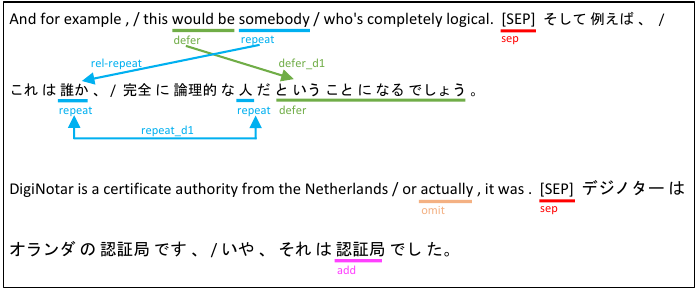}
\caption {Annotation examples. \texttt{Repeat} tags were assigned even if strings did not exactly match but referred to same entity or had same meaning.}
\label{fig:annotation-example}
\end{figure*}

We identified spans (\emph{i.e.}, words or phrases) that are \texttt{repeated}, \texttt{deferred}, or \texttt{omitted} and annotated the span tags.
In addition, we annotated \texttt{ahead}, \texttt{add}, and \texttt{error} tags for analyses of problematic translations.
The corresponding span tags in the source and CMT sentences were associated using relation tags.
Annotation examples are shown in Figure~\ref{fig:annotation-example}.

The first and second authors collaboratively annotated the first 50 examples while discussing their decisions.
Since sufficient agreement was assumed, the remainder of the data were just annotated by the first author.

\subsection{Analysis Results}

\subsubsection{Comparison among Different Translation Modes}

We compared the sentence lengths of the four datasets.
\citet{Fukuda2024-cwmt} also conducted similar comparisons based on the number of characters.
However, since variations in spelling and differences in transcription systems (\emph{e.g.}, numbers) were found, we made comparisons based on the number of words segmented by MeCab.

Table~\ref{tab:compare-sentence-length} shows that the CMT sentences were the longest, followed by offline, NAIST-SIC, and NAIST-SIC-Aligned.
These results matched those reported in \citet{Fukuda2024-cwmt}.
Long translated sentences can pose some problems.
As discussed in \citet{Fukuda2024-cwmt}, the lenght may increase the cognitive load on the listeners/readers of the translations.
In addition, longer output may cause a delay even though CMT aims to reduce it.

\begin{table}
\centering
\tabcolsep 5pt
\begin{tabular}{lrr}
\hline
\textbf{Dataset} & \textbf{Sum} & \textbf{Per Sent.$\pm$SD} \\
\hline
CMT  & 13,508 & 28.38{$\pm$18.66} \\
NAIST-SIC & 8,914 & 18.73{$\pm$12.08} \\
NAIST-SIC-Aligned & 8,072 & 16.96{$\pm$11.52} \\
Offline & 9,907 & 20.81{$\pm$12.62} \\
\hline
\end{tabular}
\caption{Comparison of number of words in different translation modes}
  \label{tab:compare-sentence-length}
\end{table}

\subsubsection{Factors that Lengthen CMT Sentences}

To reveal what factor lengthened the CMT sentences, we first analyzed them qualitatively.
Our analyses suggest that (1)~CMT sentences contain many formulaic expressions for the end of sentences as they are segmented into small chunks, and (2)~the words that are often omitted in Japanese (\emph{e.g.}, pronouns) are explicitly translated since translators were instructed to avoid \texttt{omitting} phrases in the source language, as in the following examples:
\begingroup
\addtolength\leftmargini{-0.4cm}
\begin{quote}
(En) It's when we warmed it up, / and we turned on the lights / and looked inside the box, / we saw that the piece metal / was still there in one piece. \par
\vspace{-2.0mm}
\begin{CJK}{UTF8}{min}
(CMT) {\small 私たちがそれを暖めるとき{\color{red}です}、 / 電気をつけて、 / 箱の中を見たとき{\color{red}です}、 / 私たちはその金属片を見たん{\color{red}です}、 / それはまだ一つの塊としてそこにあり{\color{red}ました}。}\\
(offline) {\small 物体を暖め明かりをつけ 箱の中を見たところ金属片はまだそこに存在してい{\color{red}ました}}
\end{CJK}
\end{quote}
\endgroup

\begingroup
\addtolength\leftmargini{-0.4cm}
\begin{quote}
(En) {\color{red}He} only has use of his eyes. \par
\vspace{-2.0mm}
\begin{CJK}{UTF8}{min}
(CMT) {\small {\color{red}彼は}目だけを使えます。}\\
(offline) {\small 目だけしか動かせません}
\end{CJK}
\end{quote}
\endgroup

In addition to the above characteristics, we observed many repetitions in particularly \emph{long} sentences.
To verify this, we further analyzed particularly \emph{long} and \emph{short} sentences, chosen based on the length ratio of the CMT sentences to the offline ones.
The \emph{long} and \emph{short} sentences were defined as those with a length ratio greater/smaller than the average $\pm$ 0.5 standard deviations (avg.=1.39, SD=0.43).
We subjectively judged whether these sentences contained many repetitions.
We also identified sentences whose offline translations were short.

Table~\ref{tab:comparison-long-short-sentences} shows that \emph{long} sentences contained more repetitions than \emph{short} ones.
The offline translation sentences were short, probably because they were originally subtitles, for which limited space was allowed.
We also quantitatively checked them using the number of assigned \texttt{repeat} tags and found that the frequency of \texttt{repetition} tags was higher in \emph{long} sentences (Table~\ref{tab:comparison-long-short-sentences}).

\begin{table}
\centering
\begin{tabular}{lrrrr}
\hline
\textbf{Type} & \textbf{\textit{N}} & \multicolumn{1}{c}{\textbf{Repeat}} & \multicolumn{1}{c}{\textbf{Repeat}} & \textbf{Short} \\
 &  & \multicolumn{1}{c}{\textbf{(subjective)}} & \multicolumn{1}{c}{\textbf{(tag)}} & \textbf{offline} \\
\hline
Long  & 131 & 54 (41.22\%) & 3.35 & 53 \\
Short & 171 & 22 (12.87\%) & 0.81 & -- \\
\hline
\end{tabular}
\caption{Comparison between \emph{long} and \emph{short} CMT sentences. Number of repeat tags is denoted per sentence.}
  \label{tab:comparison-long-short-sentences}
\end{table}

\subsubsection{Omission in SI Sentences}

To find techniques for shortening translations, we analyzed the SI sentences in NAIST-SIC.
Based on the length ratio of SI sentences to the offline ones, we defined SI sentences that might have reasonable omissions (\texttt{omission}; $0.6\leq$ ratio $<0.9$) and SI sentences that probably failed to fully convey the meaning of the source sentences (\texttt{undertranslation}; ratio $<0.6$), following the criteria in \citet{higashiyama-2023}.

Although we expected to identify some trends (\emph{e.g.}, part-of-speech) in the phrases that were omitted, we did not do so.
In addition, we found a certain number of \emph{unacceptable} translations in both categories (43.12\% and 60.00\% for \texttt{omission} and \texttt{undertranslation}, respectively).
The results suggest that human simultaneous interpreters judge the importance of phrases based on context and decide whether to translate them; some judgements are correct, and some are not.

\subsubsection{Factors that Make Monotonic Translation Difficult }

With the help of tags annotated to the source and CMT sentences, we analyzed the factors that make monotonic translation difficult.
Table~\ref{tab:compare-operation} shows the number of source phrases that were \texttt{repeated}, \texttt{deferred}, or \texttt{omitted}.
The values are based on the number of \texttt{rel-repeat}, \texttt{defer\_d\#}, and \texttt{omit} tags.
We counted the relation tags for \texttt{repeat} and \texttt{defer} because the span tags for those two operations were assigned to both the source and CMT sentences.
The results show that the translators used \texttt{repeat} most frequently, followed by \texttt{defer} and \texttt{omit}, as they were instructed (see Section~\ref{ssec:Data}).

\begin{table}
\centering
\begin{tabular}{lr}
\hline
\textbf{Operation} & \textbf{\textit{N}} \\
\hline
repeat & 301 \\
defer & 173 \\
omit & 36 \\
\hline
\end{tabular}
\caption{Comparison of number of operations used in CMT sentences}
  \label{tab:compare-operation}
\end{table}

For these phrases, we explored what makes monotonic translations difficult.
Our analyses revealed that most cases of \texttt{repeat} and \texttt{defer} were caused by particular grammatical structures.
Table~\ref{tab:syntactic-factors} lists the major structures along with their frequencies in the data and examples.
In these structures, a phrase in a chunk is typically a dependent of a phrase in the preceding chunk.
In the example of a post-modifier (Table~\ref{tab:syntactic-factors}), the relative pronoun clause is a dependent of the noun phrase \emph{a device}, which is in a preceding chunk.
When phrases with a dependency relation exist across multiple chunks, CMT is difficult because Japanese is a strongly head-final language.
The examples in Table~\ref{tab:syntactic-factors} show how human translators address these structures by repeating or deferring some phrases in subsequent chunks.

Prepositions, post-modifiers, and dependent clauses have also been identified as syntactic factors that affect interpreters' word order decisions in \citet{cai-2020-affects}.
Human interpreters find these structures challenging for SI and adopt a strategy to maintain the word order of the source language.

\begin{table*}
\centering
\small
\tabcolsep 2.5pt
\begin{tabularx}{\linewidth}{lrrX}
\hline
\textbf{Structures} & \textbf{\# repeat} & \textbf{\# defer} & \textbf{Examples} \\
\hline
Noun with a post-modifier & 88 & 12 & And now we've created {\color{red}a device} / that has absolutely no limitations.\\
 & & & \begin{CJK}{UTF8}{min} {\scriptsize さて、私たちは{\color{red}デバイス}を作り出しました、 / 全く制限のない{\color{red}もの}です。} \end{CJK} [And we've created a device / one that has absolutely no limitations] \\
Head followed by multiple dependents & 35 & 6 & ... / {\color{red}allows} for deep squats, / crawls and high agility movements. \\
 & & & \begin{CJK}{UTF8}{min} {\scriptsize ... / 深いスクワットを{\color{red}可能にし}、 / クロールや高い敏捷性の動きを{\color{red}可能にします}。} \end{CJK} [... / allows for deep squats / allows for crawls and high agility movements] \\
Dependent conjunction & 26 & 6 & ... / {\color{red}when} he's covered / in four feet of snow. \\
 & & & \begin{CJK}{UTF8}{min} {\scriptsize ... / 彼が / 四フィートの雪に覆われてしまっていた{\color{red}時には}。} \end{CJK} [he / when was covered in for feet of snow] \\
Chunk boundary before a clause & 15 & 13 & ... / you know that this {\color{red}isn't /} how it normally goes. \\
 & & & \begin{CJK}{UTF8}{min} {\scriptsize / あなたは分かるはずです、これは / 通常の進行{\color{red}ではない}ということが。} \end{CJK} [you know that this / isn't how it normally goes] \\
Chunk boundary before a preposition & 10 & 7 & ... / providing totalitarian governments with tools / {\color{red}to do this / against} their own citizens. \\
 & & & \begin{CJK}{UTF8}{min} {\scriptsize ... / 全体主義政府にツールを提供しているということです、 / {\color{red}これを行うための、} / 自国の市民に{\color{red}対してこれを行うための}ツールを。} \end{CJK} [providing totalitarian governments with tools / to do this / tools to do this against their own citizens] \\
\hline
\end{tabularx}
\caption{Syntactic factors that prevent monotonic translations. Cases involving multiple structures were classified separately as \emph{compound factors}.}
  \label{tab:syntactic-factors}
\end{table*}

In addition, we observed that inappropriate segmentation was addressed by \texttt{repeating} and \texttt{deferring} the phrases.
Most inappropriate segmentation was found in phrasal verbs, verbal gerunds, and to-infinitives.

In the SI data, we also found that human interpreters \texttt{repeat} phrases to maintain the word order of the source language.
For example, in the example in Table~\ref{tab:comparison-of-translation-mode}, a noun modified by a preposition phrase is \texttt{repeated}:
\begingroup
\addtolength\leftmargini{-0.4cm}
\begin{quote}
(En) ... / froze the Swiss bank {\color{red}account} / of Mr. Sam Jain right here, / ... \par
\vspace{-2.0mm}
\begin{CJK}{UTF8}{min}
(SI) {\small ... / スイスの銀行の{\color{red}口座}を凍結しました。 / サムジェインの{\color{red}もの}です。 / ...}\\
\end{CJK}
[... / froze the Swiss bank account / it is Sam Jain's one / ...]
\end{quote}
\endgroup

\noindent
In addition, \citet{okamura-2023} reported that the order of the chunks is shuffled about once on average in an SI sentence.
These things suggest that human interpreters address the word order differences that make monotonic translation difficult by \texttt{repeating} and \texttt{deferring} some phrases.

\section{Evaluation Using CMT sentences}

To investigate the impact of using CMT sentences for evaluating translation quality, we evaluated the output from existing ST and simulST models using the NAIST English-to-Japanese Chunk-wise Monotonic Translation Evaluation Dataset as well as existing test sets.

\subsection{Data}
\label{ssec:eval-exp-data}

We used the following four datasets as references for the automatic evaluation metrics:
\begin{itemize}
    \setlength{\itemsep}{-3pt}
    \item \textbf{n-cmt}: CMT sentences from the NAIST English-to-Japanese Chunk-wise Monotonic Translation Evaluation Dataset
    \item \textbf{si\_hum}: SI sentences from NAIST-SIC, manually aligned to the source speech
    \item \textbf{si\_auto}: SI sentences from NAIST-SIC-Aligned, aligned automatically
    \item \textbf{offline}: offline translation sentences from the subtitles of TED talks.
\end{itemize}

\noindent
Because \texttt{si\_auto} was created by applying automatic alignment and filtering techniques to SI sentences in \texttt{si\_hum}, it may contain alignment errors.
In addition, SI sentences in \texttt{si\_auto} tend to be shorter than those in \texttt{si\_hum} (see Table~\ref{tab:compare-sentence-length}).
Therefore, we used the two SI-based datasets for our evaluation.

\subsection{Speech Translation Models}
\label{ssec:st-model}

We used three existing models (\emph{i.e.}, one ST and two simulST models):
\begin{itemize}
    \setlength{\itemsep}{-3pt}
    \item \textbf{ST\_offline}: an ST model trained on offline data \cite{fukuda-etal-2023-naist}
    \item  \textbf{simulST\_offline}: a simulST model trained on offline data \cite{ko-etal-2023-tagged}
    \item  \textbf{simulST\_si\_offline}: a simulST model trained on both offline and SI data \cite{ko-etal-2023-tagged}.
\end{itemize}
All the models were built by connecting two pre-trained models, HuBERT-Large \cite{hsu2021hubert} for their speech encoder and the decoder of mBART50 \cite{tang2020multilingual} for their text decoder.
The encoder and decoder were connected by Inter-connection \cite{nishikawa23_interspeech} and a length adapter \cite{tsiamas-etal-2022-pretrained}.
Both SimulST models used bilingual prefix pairs extracted using Bilingual Prefix Alignment \cite{kano-etal-2022-simultaneous} for the model training and employed a decoding policy called local agreement \cite{liu20s_interspeech}.
For ST\_offline, we used a model with checkpoint averaging (\texttt{Inter-connection + Ckpt Ave.} in \citet{fukuda-etal-2023-naist}).
For simulST\_offline and simulST\_si\_offline, we used the models that satisfy the task requirement of the simultaneous track in the IWSLT 2023 Evaluation Campaign\footnote{\url{https://iwslt.org/2023/simultaneous}}, latency measured by Average Lagging \cite{ma-etal-2019-stacl} $\leq 2$ seconds (\texttt{Offline FT} and \texttt{Mixed FT + Style} in \citet{ko-etal-2023-tagged}, respectively).


\begin{table*}
\centering
\small
\tabcolsep 1.7pt
\begin{tabular}{l|rrrr|rrrr|rrrr}
\hline
\textbf{Model} & \multicolumn{3}{l}{\textbf{BLEU}} & & \multicolumn{3}{l}{\textbf{BLEURT}} & & \multicolumn{4}{l}{\textbf{COMET}} \\
 & n-cmt & si\_hum & si\_auto & offline & n-cmt & si\_hum & si\_auto & offline & n-cmt & si\_hum & si\_auto & offline \\
\hline
ST\_offline & 14.487 \, & 8.856 \, & 8.637 \, & 17.775 \, & 0.553 \;\; & 0.447 \;\; & 0.414 \;\; & \textbf{0.538} \;\; & \textbf{0.838} \;\; & \textbf{0.797} \;\; & \textbf{0.781}$^{*1}$ & \textbf{0.833} \;\; \\
simulST\_offline & 15.406$^{\dag}$ & 8.446$^{\dag}$ & 7.773$^{\dag}$ & \textbf{17.907} \, & 0.556 \;\; & 0.442 \;\; & 0.406 \;\; & 0.531 \;\; & 0.826 \;\; & 0.780 \;\; & 0.763 \;\; & 0.821 \;\; \\
simulST\_si\_offline & \textbf{15.982}$^{\dag}$ & \textbf{12.031}$^{\dag}$ & \textbf{11.020}$^{\dag}$ & 13.191$^{\dag}$ & \textbf{0.567} \;\; & \textbf{0.493}$^{*1}$ & \textbf{0.460}$^{*1}$ & 0.519 \;\; & 0.807$^{*2}$ & 0.774$^{*3}$ & 0.761 \;\; & 0.789$^{*2}$ \\
\hline
\multicolumn{13}{l}{} \\
\end{tabular}
\begin{tabular}{l|rrrr|rrrr|rrrr}
\hline
\textbf{Model} & \multicolumn{3}{l}{\textbf{BERTScore (Pre.)}} & & \multicolumn{3}{l}{\textbf{BERTScore (Rec.)}} & & \multicolumn{4}{l}{\textbf{BERTScore (F1)}} \\
 & n-cmt & si\_hum & si\_auto & offline & n-cmt & si\_hum & si\_auto & offline & n-cmt & si\_hum & si\_auto & offline \\
\hline
ST\_offline & \,0.801 \;\; & 0.735 \;\; & 0.722 \;\; & \textbf{0.789} \;\; & 0.769 \;\; & 0.739 \;\; & 0.735 \;\; & \textbf{0.788} \;\; & 0.784 \;\; & 0.737 \;\; & 0.728 \;\; & \textbf{0.788} \;\; \\
simulST\_offline & \,0.799 \;\; & 0.730 \;\; & 0.717 \;\; & 0.783 \;\; & 0.770 \;\; & 0.738 \;\; & 0.734 \;\; & 0.786 \;\; & 0.783 \;\; & 0.734 \;\; & 0.725 \;\; & 0.784 \;\; \\
simulST\_si\_offline & \,\textbf{0.817}$^{*1}$ & \textbf{0.764}$^{*1}$ & \textbf{0.746}$^{*1}$ & 0.759$^{*2}$ & \textbf{0.784}$^{*1}$ & \textbf{0.766}$^{*1}$ & \textbf{0.760}$^{*1}$ & 0.757$^{*2}$ & \textbf{0.800}$^{*1}$ & \textbf{0.764}$^{*1}$ & \textbf{0.752}$^{*1}$ & 0.757$^{*2}$ \\
\hline
\end{tabular}
\caption{Results of quality evaluation metrics across ST and simulST models. $^{\dag}$: significantly different from ST\_offline.  $*1$: significantly higher than other two. $*2$ significantly lower than other two. $*3$ significantly lower than ST\_offline. Significance threshold was set to $p<.05$ for all tests.}
  \label{tab:results-evaluation}
\end{table*}

\subsection{Metrics}

We evaluated the translation quality of the output from the ST and simulST models (see Section~\ref{ssec:st-model}) using BLEU\footnote{BLEU was calculated using sacreBLEU. \cite{post-2018-call} \url{https://github.com/mjpost/sacrebleu}}, BLEURT\footnote{\url{https://github.com/google-research/bleurt}} \cite{sellam-etal-2020-bleurt}, COMET\footnote{\url{https://github.com/Unbabel/COMET}} \cite{rei-etal-2020-comet}, and BERTScore\footnote{\url{https://github.com/Tiiiger/bert_score}} \cite{zhang-2020-BERTScore}.
BERTScore was calculated using \texttt{bert-base-multilingual-cased}.
We used the four datasets described in Section~\ref{ssec:eval-exp-data} as references.

\subsection{Evaluation Results}

Table~\ref{tab:results-evaluation} shows the results of the quality evaluation metrics across the ST and simulST models.
For the BLEU scores, we conducted paired significance tests using paired bootstrap resampling \cite{koehn-2004-statistical}.
We specified \texttt{ST\_offline} as the baseline for the significance tests.
For the other scores, we conducted a one-way ANOVA, followed by Tukey's multiple comparisons test.

When the translation quality was evaluated using BLEU with \texttt{n-cmt} as the reference, \texttt{simulST\_si\_offline} achieved the highest score.
On the SI-based test sets (\emph{i.e.}, \texttt{si\_hum} and \texttt{si\_auto}), \texttt{simulST\_si\_offline} also had the highest score.
On the offline-based test set, in contrast, the models trained on only offline data achieved much higher scores than \texttt{simulST\_si\_offline}.
The same tendencies were observed in BLEURT and BERTScore.
These results suggest that the models trained on both SI and offline data generated more SI-like translations, and such models perhaps should be evaluated using a reference closer to SI sentences.
In addition, using an offline-based test set might underestimate the performance of models trained on both SI and offline data.

Comparing \texttt{n-cmt}, \texttt{si\_hum}, and \texttt{si\_auto}, the scores were highest for \texttt{n-cmt}, followed by \texttt{si\_hum} and \texttt{si\_auto} on all the metrics and models.
Because \texttt{si\_hum} is based on SI sentences generated by human simultaneous interpreters, some content in the source speech might be omitted or inadequately translated (under-translation).
SI sentences in \texttt{si\_auto}, which were automatically created based on human SI sentences, might contain less source speech content than those in \texttt{si\_hum} due to the alignment and filtering techniques applied (see \citealp{zhao-2024-naist-sic-aligned}).
In fact, BERTScore precision was higher than recall on \texttt{n-cmt}, in which there were almost no omissions, while recall was higher than precision on \texttt{si\_auto} and precision and recall were almost equal on \texttt{si\_hum}.
These results indicate the possibility that the existing SI-based test sets \cite{ko-etal-2023-tagged,zhao-2024-naist-sic-aligned} underestimate the model performance.

However, the COMET results were different from those on the other metrics (Table~\ref{tab:results-evaluation}).
On all four test sets, \texttt{ST\_offline} achieved the highest score, followed by \texttt{simulST\_offline} and \texttt{simulST\_si\_offline}.
One possible reason is that COMET uses source sentences to calculate its scores.

To examine the impact of the source sentences, we also calculated a reference-free COMET-QE using \texttt{wmt22-cometkiwi-da} and got similar results (0.813, 0.798, and 0.766 for \texttt{ST\_offline}, \texttt{simulST\_offline}, and \texttt{simulST\_si\_offline}, respectively).
We further calculated COMET-QE for \texttt{n-cmt} and \texttt{offline}, regarding them as oracle data, and found that \texttt{n-cmt} had a higher score than \texttt{offline} (\texttt{n-cmt}: 0.832, \texttt{offline}: 0.812).
Because some translation sentences in \texttt{offline} are under-translated, these results suggest that the COMET scores tend to become high when more content in the source sentences is covered in the target sentences.
This feature does not fit the nature of SI, where human interpreters use sophisticated strategies (see \citealp{he-etal-2016-interpretese, cai-2020-affects}, for example).
We need to carefully interpret COMET scores when we use them for evaluating simulST models.

\section{Conclusion}

This paper focused on monotonic translations in English-Japanese SI.
Our analyses revealed some grammatical structures that make monotonic translations difficult and that human interpreters/translators address these challenges by repeating or deferring some phrases in source language in the subsequent chunks.
The grammatical structures that might cause delays would be useful information for developing segmentation or decoding policies for simultaneous machine translation systems.
One possible direction would be predicting whether a phrase in a chunk is the head of a phrase in subsequent chunks.

We also evaluated the output from the existing ST and simulST models on the NAIST English-to-Japanese Chunk-wise Monotonic Translation Evaluation Dataset as well as on existing SI-based and offline-based test sets.
The BLEU, BLEURT, and BERTScore results supported using CMT sentences for evaluating simulST models trained using SI data, although the results with COMET were different.
Further analysis across various evaluation metrics is necessary.
Analyzing how the source and target sentences are aligned monotonically on different types of translations (\emph{e.g.}, \citealp{han-etal-2021-monotonic}) would also be useful.

This paper investigated the impact of using CMT sentences for evaluation purposes.
A future study would involve using monotonic translation sentences for developing simulST models \cite{sakai-2024-simultaneous}\footnote{Published around the same time as the submission of this paper.}.
It could potentially address the problem that simulST models trained using SI sentences suffered from under-translation \cite{ko-etal-2023-tagged}.
However, CMT sentences tend to be long.
Investigating the trade-offs between longer CMT sentences and the potential cognitive load on listeners/readers
might provide further insights.

\section*{Acknowledgments}
A part of this work was supported by JSPS KAKENHI Grant Numbers JP21H05054 and by JST, the establishment of university fellowships towards the creation of science technology innovation, Grant Number JPMJFS2137.

\bibliography{anthology,custom}




\end{document}